\documentclass[a4paper, 10pt, conference]{ieeeconf}      

\IEEEoverridecommandlockouts                              

\usepackage{graphicx}
\usepackage{booktabs}
\usepackage{pifont}
\usepackage{xcolor}
\newcommand{\cmark}{\textcolor{green!80!black}{\ding{51}}}
\newcommand{\xmark}{\textcolor{red}{\ding{55}}}

\usepackage{hyperref}
\hypersetup{
    colorlinks=true,
    linkcolor=blue,
    filecolor=magenta,      
    urlcolor=magenta,
    pdftitle={Overleaf Example},
    pdfpagemode=FullScreen,
    }
\usepackage{float}
\usepackage{makecell}
\definecolor{cvprblue}{rgb}{0.21,0.49,0.74}
\usepackage{framed}
\usepackage{tikz} 
\usepackage{adjustbox}
\usepackage{multirow}
\usepackage{array}
\newcommand{\minus}{\scalebox{0.75}[1.0]{$-$}}

\newcommand\RaiseImage[2][scale=1]{%
  \raisebox{-0.5\totalheight}{\includegraphics[#1]{#2}}}
  
\newcommand{\mybox}[4]{
    \begin{figure}[t]
    \label{prompt:instruction}
        \centering
    \begin{tikzpicture}
        \node[anchor=text,text width=\columnwidth-1.2cm, draw, rounded corners, line width=1pt, fill=#3, inner sep=5mm] (big) {\\#4};
        \node[draw, rounded corners, line width=.5pt, fill=#2, anchor=west, xshift=5mm] (small) at (big.north west) {#1};
    \end{tikzpicture}
    \caption{\textbf{Prompt for instructing of safety measurement}} \label{fig:prompt_instruction}
    \end{figure}
}

\newcommand{\myboxcriteria}[4]{
    \begin{figure}[t]
        \centering
    \begin{tikzpicture}
        \node[anchor=text,text width=\columnwidth-1.2cm, draw, rounded corners, line width=1pt, fill=#3, inner sep=5mm] (big) {\\#4};
        \node[draw, rounded corners, line width=.5pt, fill=#2, anchor=west, xshift=5mm] (small) at (big.north west) {#1};
    \end{tikzpicture}
 \caption{\textbf{Prompt for evaluating safety score criteria}} 
    \label{prompt:criteria}
    \end{figure}
}

\newcommand{\myboxcot}[4]{
    \begin{figure}[h]
        \centering
    \begin{tikzpicture}
        \node[anchor=text,text width=\columnwidth-1.2cm, draw, rounded corners, line width=1pt, fill=#3, inner sep=5mm] (big) {\\#4};
        \node[draw, rounded corners, line width=.5pt, fill=#2, anchor=west, xshift=5mm] (small) at (big.north west) {#1};
    \end{tikzpicture}
     \caption{\textbf{Prompt for Auto Chain-of-Thought}} 
    \label{prompt:cot}
    \end{figure}
}

\definecolor{myblue}{RGB}{218, 235, 253}
\definecolor{mygreen}{RGB}{214, 226, 211}
\definecolor{myyellow}{RGB}{245, 229, 164}
\definecolor{myred}{RGB}{236, 212, 209}
\definecolor{mygray}{RGB}{194, 192, 182}
\definecolor{tablegray}{rgb}{0.2,0.2,0.2}

\usepackage{threeparttable}

\title{\LARGE \bf
Is it safe to cross? Interpretable Risk Assessment with GPT-4V for
Safety-Aware Street Crossing}

\author{Hochul Hwang$^{1}$, Sunjae Kwon$^{1}$, Yekyung Kim$^{1}$, and Donghyun Kim$^{1}$
\thanks{$^{1}$ Manning College of Information and Computer Sciences, University of Massachusetts Amherst, 140 Governors Dr, Amherst, MA 01002, U.S. {\tt\small\{hochulhwang, sunjaekwon, yekyungkim, donghyunkim\}@cs.umass.edu}}
}

\begin{document}

\maketitle

\thispagestyle{empty}
\pagestyle{empty}

\begin{abstract}

Safely navigating street intersections is a complex challenge for blind and low-vision individuals, as it requires a nuanced understanding of the surrounding context -- a task heavily reliant on visual cues. Traditional methods for assisting in this decision-making process often fall short, lacking the ability to provide a comprehensive scene analysis and safety level. This paper introduces an innovative approach that leverages vision-language models (VLMs) to interpret complex street crossing scenes, offering a potential advancement over conventional traffic signal recognition techniques. By generating a safety score and scene description in natural language, our method supports safe decision-making for blind and low-vision individuals. We collected crosswalk intersection data that contains multiview egocentric images captured by a quadruped robot and annotated the images with corresponding safety scores based on our predefined safety score categorization. Grounded on the visual knowledge, extracted from images and text prompts, we evaluate a VLM for safety score prediction and scene description. Our findings highlight the reasoning and safety score prediction capabilities of the VLM, activated by various prompts, as a pathway to developing a trustworthy system, crucial for applications requiring reliable decision-making support.

\end{abstract}
\maketitle

\section{INTRODUCTION}
\label{sec:intro}

Safety-aware scene understanding and risk assessment are critical for effective robot navigation, particularly in safety-critical contexts such as guiding blind or low-vision (BLV) individuals to cross a street. While strides in autonomous vehicle technology have led to the release of publicly available large-scale datasets for training deep learning models~\cite{geiger2012we, cordts2016cityscapes, yu2020bdd100k}, their direct application poses challenges. This stems from the inherent differences between road and sidewalk environments~\cite{zhang2023towards}. Moreover, such vehicle-centric datasets often miss out on elements vital for blind navigation, such as tactile pavings.

Our research introduces a novel framework for assisting BLV individuals in safely navigating crosswalk intersections. Unlike traditional approaches, primarily focused on traffic light signal recognition, we leverage a vision-language model (VLM) to enhance contextual scene understanding. Recently, there has been research enabling zero-shot transfer for various computer vision tasks, including classification~\cite{radford2021learning}, segmentation~\cite{kirillov2023segment}, and multimodal reasoning~\cite{liu2023visual}, utilizing transformers~\cite{vaswani2017attention, dosovitskiy2020image} and large language models (LLMs) such as ChatGPT and GPT-4~\cite{touvron2023llama}. Here, we utilize a vision-language model for a more comprehensive and nuanced understanding of safe street crossing. This advancement potentially enables the provision of safety-critical information in natural language, significantly enhancing the decision-making process for BLV individuals.

The essence of our approach lies in the collection and usage of a custom outdoor street crossing dataset, uniquely gathered via the Unitree Go1 quadruped robot. This dataset, annotated with predefined safety scores, serves as a cornerstone in the enhancement of prompts for VLMs. By integrating \textit{visual knowledge} -- encompassing object detection bounding boxes, segmentation masks, and optical flow -- with textual information, we explore the capability of the VLM, with a particular focus on GPT-4V\footnote{https://platform.openai.com/docs/guides/vision}. This investigation aims to evaluate the model's proficiency in accurately interpreting the intricacies of crosswalk environments, thereby informing decisions for safe street crossing. 

Our findings highlight the effectiveness of the VLM and its potential for making safety-aware  decisions, advancing the domain of autonomous robot navigation in pedestrian environments. This contribution is particularly geared towards developing a robotic mobility aid for BLV individuals, offering implications for enhancing their autonomy and mobility. We summarize our contributions as follows:

\begin{itemize}
    \item Data collection of multiview egocentric images in a crosswalk environment for street crossing.
    \item Safety score categorization on egocentric images.
    \item Evaluation of GPT-4V for safety score prediction and scene description using visual and text prompts.
\end{itemize}
\section{RELATED WORK}
\label{sec:related_work}

\begin{figure*}[t]
    \label{fig:framework}
    \centering
    \includegraphics[width=\linewidth]{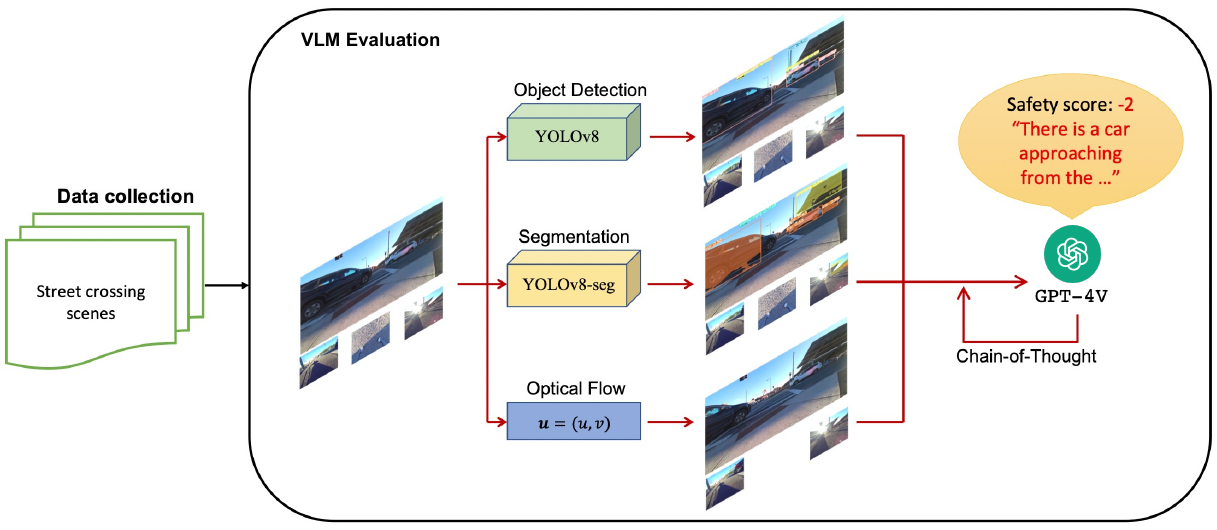}
    \caption{\textbf{Illustrative concept of VLM evaluation pipeline for safety-aware street crossing.} Visual knowledge of object detection bounding boxes, segmentation masks, and optical flow is extracted from the robot's multiview egocentric images. This is then provided to the VLM along with the text prompts. The VLM outputs the safety score and scene description.}
    \label{fig:f1}
\end{figure*}

\subsection{Assistive Technology for Blind Street Crossing} 
BLV people primarily rely on their residual senses, such as hearing traffic sounds to decide when to cross streets. However, these methods do not guarantee accurate alignment with pedestrian crossings~\cite{guth1989tests}. Crossing streets safely poses a significant challenge for blind individuals, who rely primarily on auditory cues~\cite{hassan2012normally}. Studies reveal that blind pedestrians face a higher risk and typically encounter significant delays compared to sighted pedestrians, particularly in complex traffic settings such as roundabouts~\cite{ashmead2005street}. 

Recognizing safety is paramount for street crossing, yet it may be equally crucial to provide the robot's user with a detailed scene description. Unlike traditional travel aids such as white canes and guide dogs, which do not provide feedback to the user, recent studies explore the potential of scene description through advancements in computer vision~\cite{ichikawa2022voice, liu2023dragon, guerreiro2019cabot}. Hoogsteen et al.~\cite{hoogsteen2022beyond} investigate the factors that influence the importance of describing specific objects in the environment and explore effective methods for communicating these descriptions. Their study also compares the preferences of early and late blind participants regarding object descriptions. More recent vision assistive mobile applications such as OKO~\cite{oko} and Be My Eyes~\cite{bme} show their practical applications in conveying useful information to the user. Such research aligns with the overarching theme of our paper, which focuses on integrating vision language models in robots to function as mobility aids, moving beyond the capabilities of traditional travel aids.

\subsection{Vision Language Models} 
With the advent of ChatGPT~\cite{ray2023chatgpt}, generative large language models (LLMs) have been increasingly applied to complex inferential tasks beyond the reach of natural language processing models. These tasks include solving mathematical problems~\cite{poola2023guiding}, definition modeling~\cite{kwon-etal-2023-vision}, and navigating domains that require expert knowledge (e.g., medical and financial sectors~\cite{nori2023capabilities, liang2023breaking}). In particular, models such as GPT-4 have demonstrated superior annotation quality compared to untrained human annotators (e.g., Amazon Mechanical Turk~\cite{gilardi2023chatgpt}). This opens up the possibility of using LLMs as an evaluator, offering a cost-effective alternative to human evaluations with enhanced instruction-following capabilities \cite{Dubois2023AlpacaFarmAS, Zheng2023JudgingLW, Min2023FActScoreFA, Liu2023GEvalNE}.

Current research on LLMs is expanding their application scope beyond textual analysis to include other modalities such as vision and speech~\cite{zhu2023minigpt, wu2023next,liu2023visual,chen2023minigpt}. We extend this exploration by incorporating the visual data recognition capabilities of VLMs into the annotation process. Demonstrating exceptional capability, our study employs GPT-4V to evaluate the safety of crossing streets. This approach uniquely integrates various visual knowledge with text prompts, laying the groundwork for informed and safety-aware decision-making in street crossing scenarios.

\section{METHOD}
\label{sec:method}

\begin{table*}[!ht]
\centering
\begin{tabular}{@{}l|cccccc@{}}
\toprule
\textbf{} & \textbf{Totally unsafe} & 
\multicolumn{2}{c}{\textbf{Partially unsafe}}
& \textbf{Keep caution} & \textbf{Partially safe} & \textbf{Totally safe} 
\\ \midrule
\rotatebox[origin=c]{90}{Image} 
&
\RaiseImage[width=2.7cm]{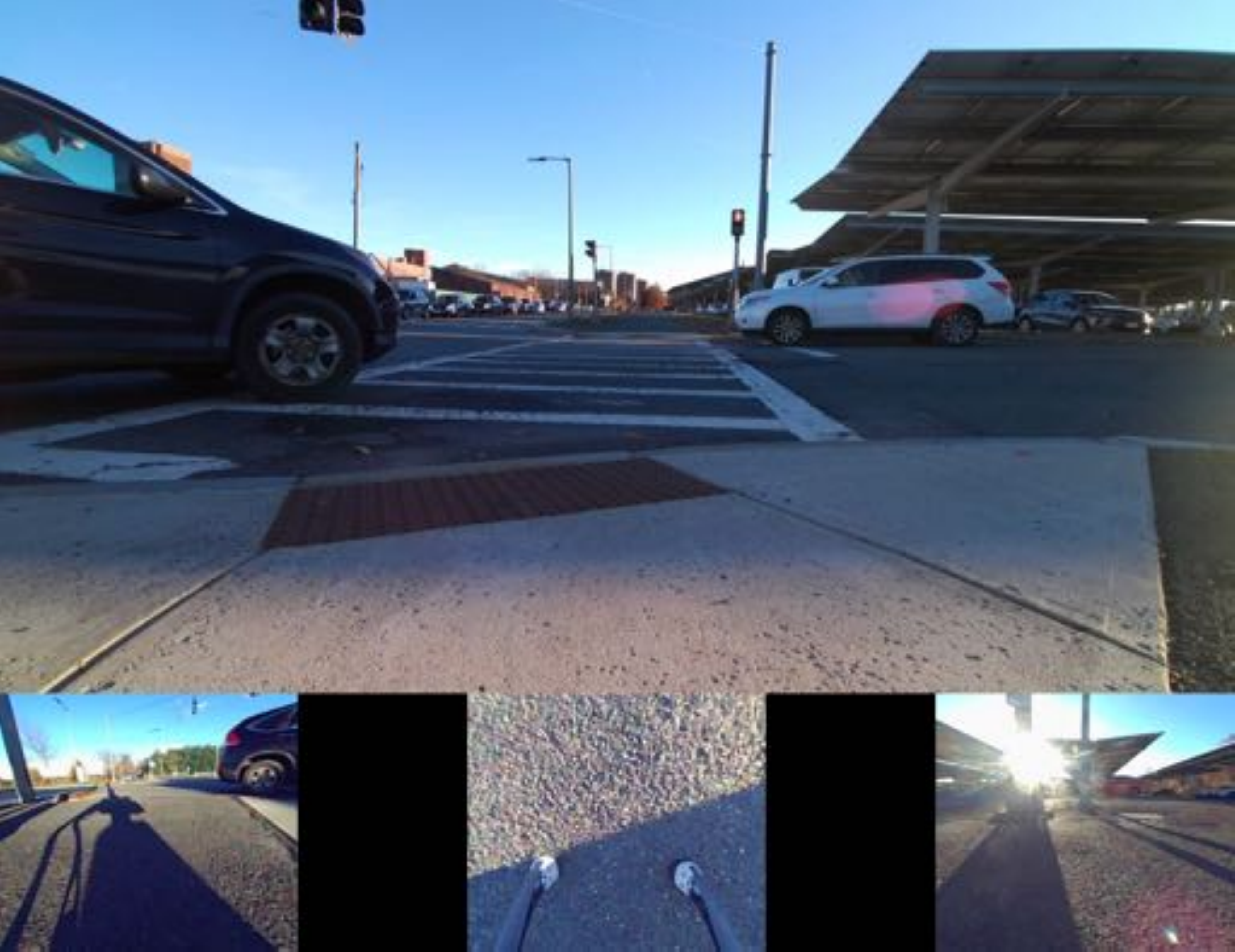}
& &
\RaiseImage[width=2.7cm]{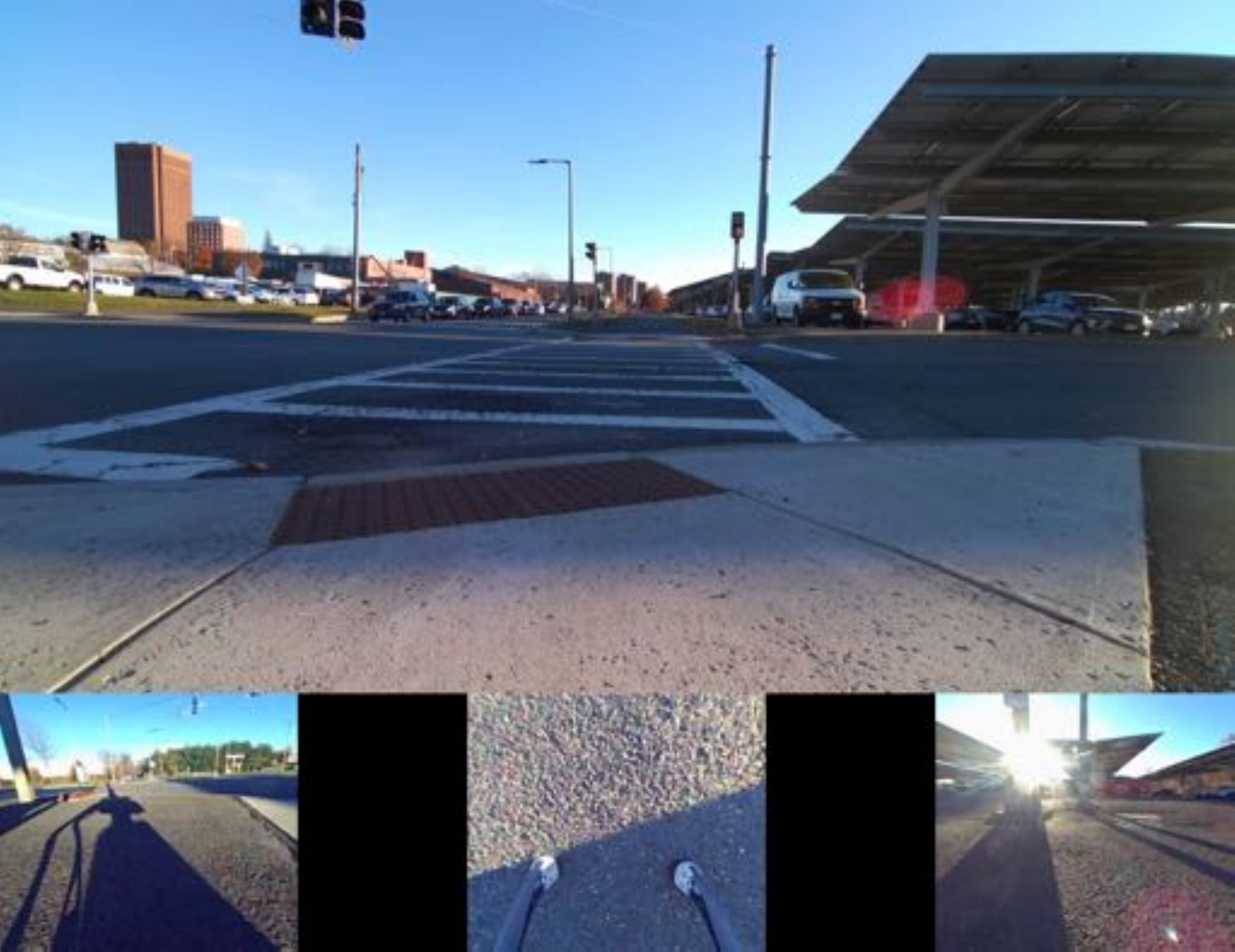}
&
\RaiseImage[width=2.7cm]{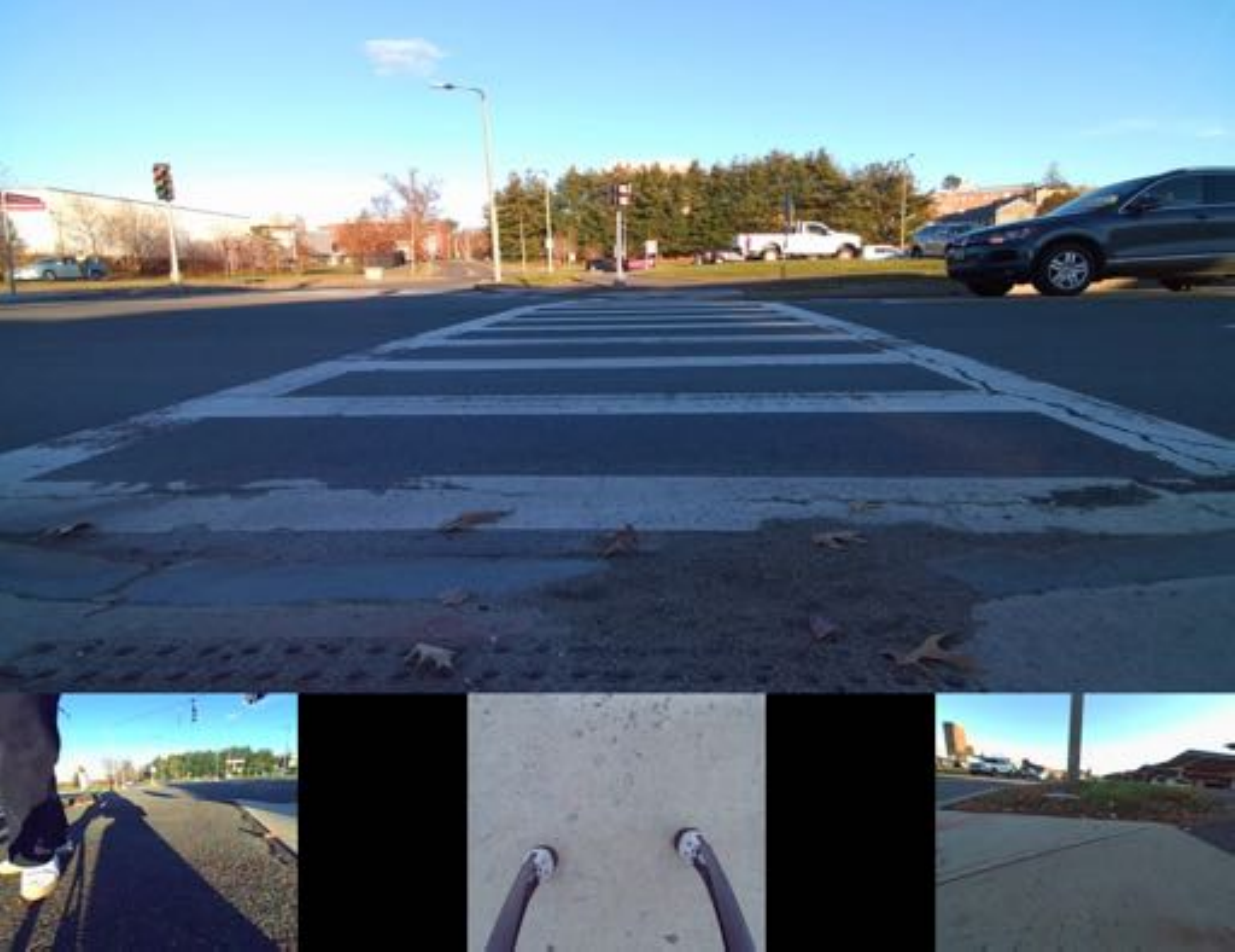}
&
\RaiseImage[width=2.7cm]{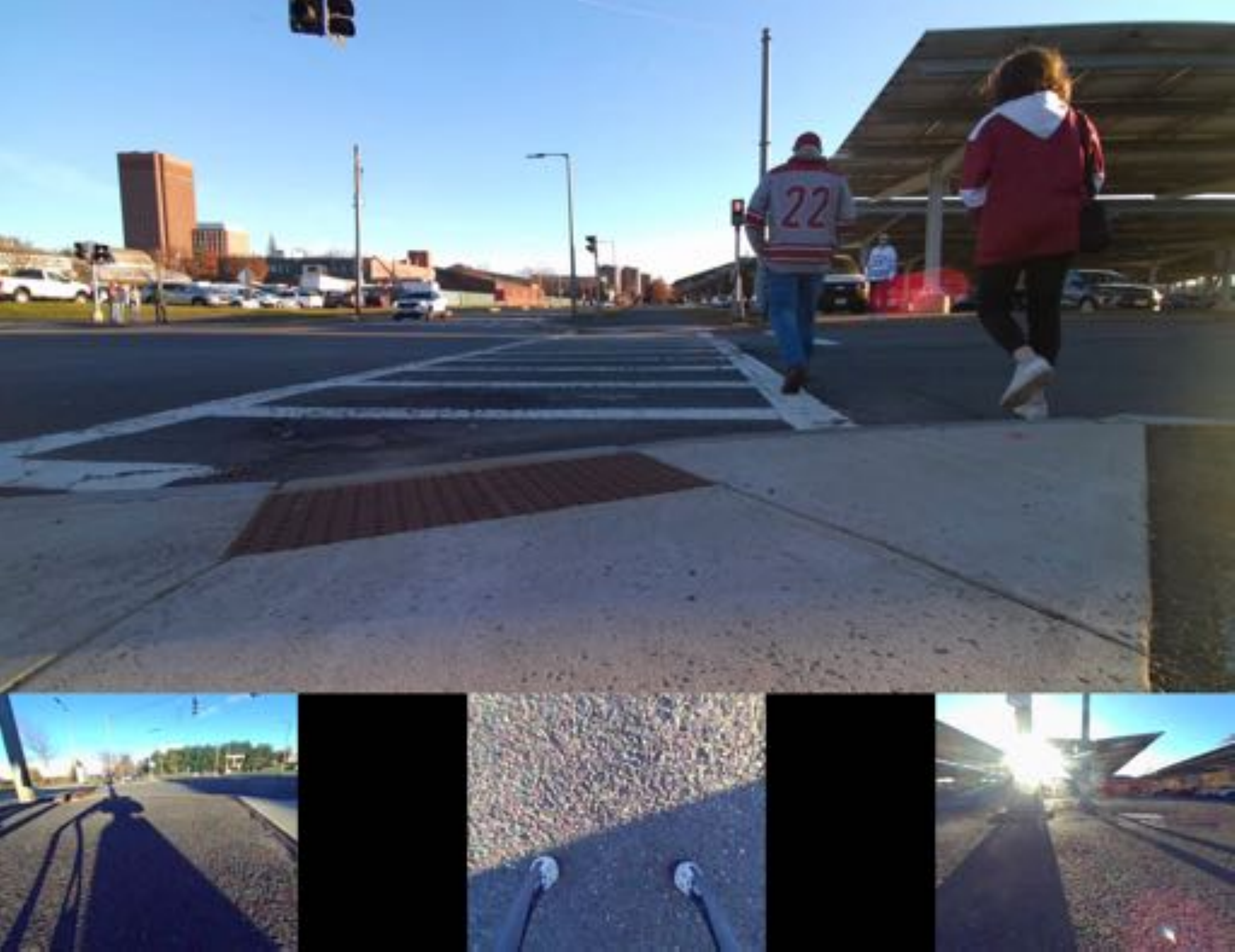}
&
\RaiseImage[width=2.7cm]{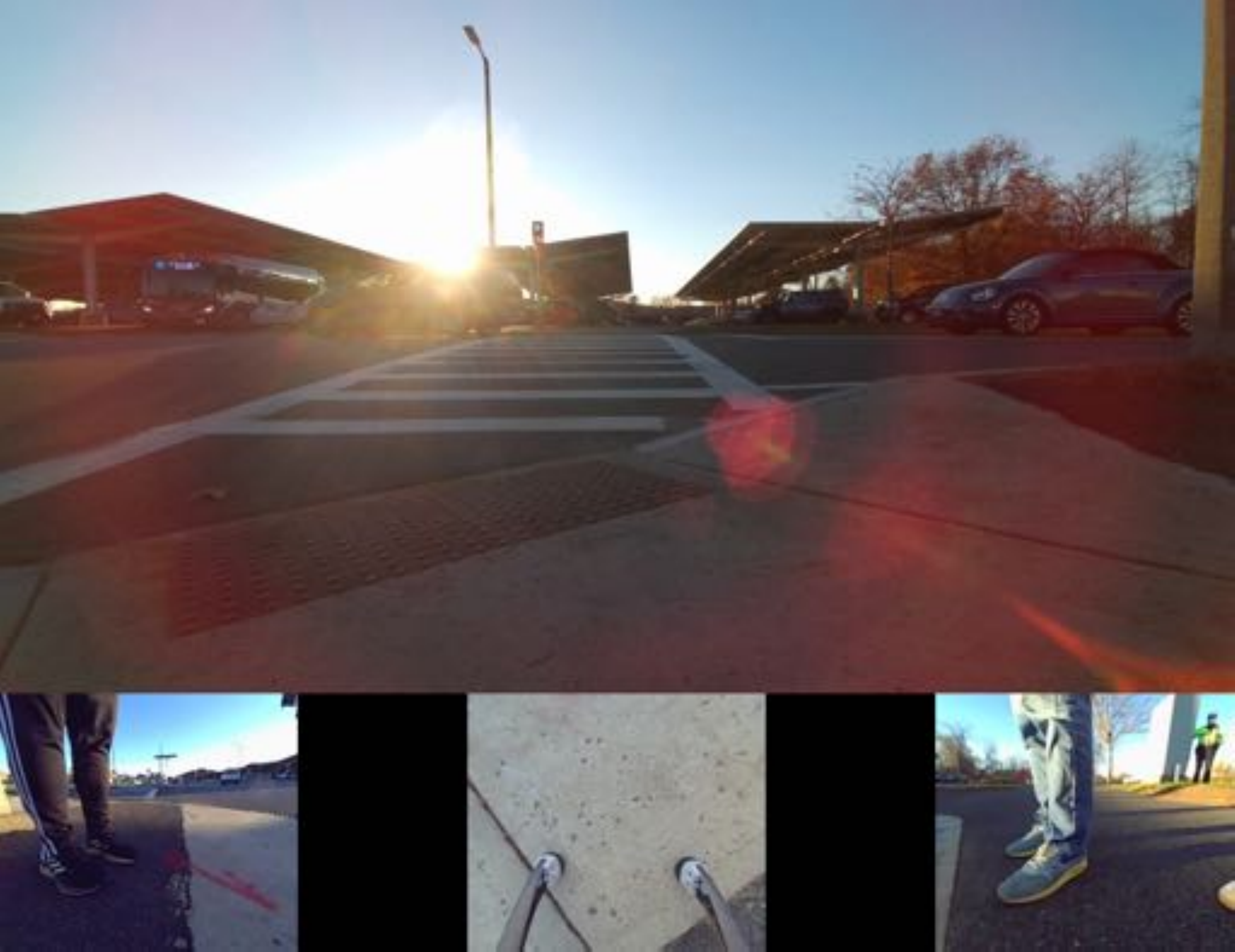}

\\
\bottomrule
\end{tabular}
\caption{Visualization of egocentric images from each safety level.} 
\label{tab:example_image}
\end{table*}

\begin{table*}
  \centering
  \begin{tabular}{@{}lccccc@{}}
    \toprule
    Safety Score & Safe to cross & Moving car & Parallel traffic light & Pedestrian signal & Crossing pedestrian \\
    \midrule
    Totally unsafe (-2) &\xmark 
    & yes & all & all & all
    \\
    Partially unsafe (-1)&\xmark 
    & no, not visible & all & stop & all
    \\ 
    Keep caution (0)  &-
    & no, not visible & \colorbox{myred}{red}, \colorbox{myyellow}{yellow}, not visible & go & no, not visible
    \\ 
    \multirow{2}{*}{Partially safe (1)} & \multirow{2}{*}{\cmark} 
    & no, not visible & \colorbox{myred}{red}, \colorbox{myyellow}{yellow}, not visible & go, not visible & yes
    \\ 
    && no, not visible & \colorbox{mygreen}{green} & go & no, not visible
    \\ 
    Totally safe (2) & \cmark
    & no, not visible & \colorbox{mygreen}{green} & go, not visible & yes
    \\
    
 \bottomrule
  \end{tabular}
    \caption{Safety score categorization.}
  \label{tab:safety}
\end{table*}


  
    

In this section, we introduce our evaluation pipeline for risk assessment in a street crossing scenario, as illustrated in Fig.~\ref{fig:f1}. Our framework consists of two major parts: visual knowledge extraction and evaluation with VLM using visual and text prompts. 


\begin{figure}
    \centering
    \includegraphics[width=\linewidth]{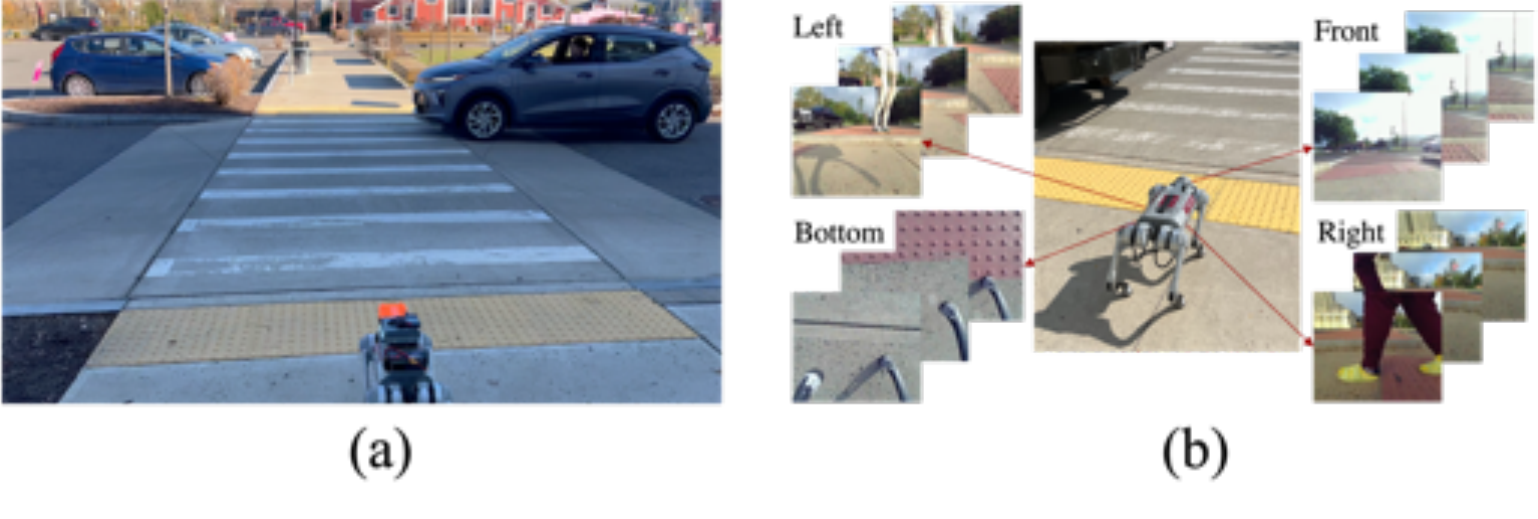}        
    \vspace{-5mm}
    \caption{\textbf{Data collection using a quadruped robot.} (a) The remote controlled robot collected RGB and depth data in crosswalk settings. One of the researchers is driving the car to simulate an unsafe scenario for street crossing. (b) Various onboard sensors enable egocentric multiview data.}
    \label{fig:f3_collection}
\end{figure}

\subsection{Visual Knowledge as Prompts}
\subsubsection{Crosswalk data collection}
To evaluate the capabilities of the vision-language model, we collected multiview egocentric images of real-world crosswalk scenarios including situations that are not safe for pedestrians to cross the street as illustrated in Fig.~\ref{fig:f3_collection}. All researchers participated in data collection while one researcher remotely controlled the robot and another drove a vehicle to create realistic scenarios. We utilized three onboard stereo cameras to capture the right, left, and bottom viewpoints of the robot, and one Azure Kinect camera to capture the front viewpoint. While depth information was also captured, our study focused exclusively on the RGB data. Data collection occurred at various times throughout the day to ensure a diverse representation of lighting and environmental conditions.

\subsubsection{Safety score categorization}
Safety score is categorized into five levels (see Table~\ref{tab:example_image}) to be used as robot feedback. We employed a detailed approach to label image-safety score pairs, as outlined in Table~\ref{tab:safety}. This labeling was based on predefined rules, taking into account various critical factors necessary for ensuring safe street crossing. Specifically, the safety level was assigned based on (1) presence of moving vehicles toward the crosswalk: \{yes, no, not visible\}; (2) status of the parallel traffic light: \{\colorbox{myred}{red}, \colorbox{myyellow}{yellow}, \colorbox{mygreen}{green}, not visible\}; (3) state of the pedestrian signal: \{go, stop, not visible\}; (4) and presence of pedestrians in the crosswalk: \{yes, no, not visible\}. 

Contrary to previous studies that categorized street crossing as a binary decision~\cite{radwan2020multimodal}, our method introduces a nuanced safety score, $s \in \{\minus 2, \minus 1, 0, 1, 2\}$, with the highest value indicating the safest situation for street crossing. To ensure accuracy and consistency in the labeling, our team meticulously reviewed each piece of data.

\begin{figure}
    \centering
    \includegraphics[width=0.9\linewidth]{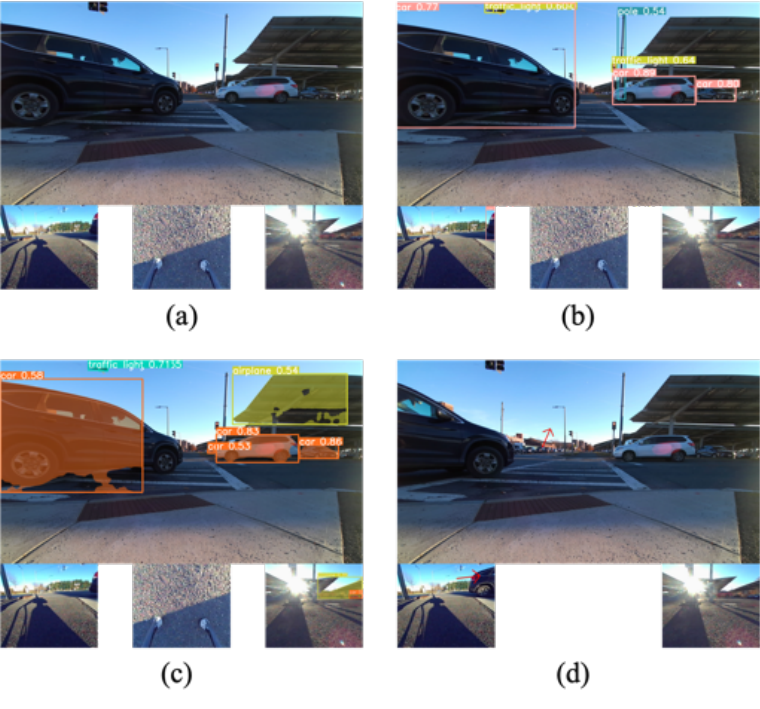}
    \vspace{-1mm}
    \caption{\textbf{Visualization of visual knowledge.} (a) Raw multiview images contain the egocentric front (top patch), left (bottom left patch), bottom (bottom center patch), and right (bottom right patch) viewpoints of the robot. (b) Bounding boxes are added using an object detection algorithm. (c) Segmentation masks are added using an instance segmentation algorithm. (d) The average optical flow for each viewpoint (except for the bottom) is represented by a red arrow.}
    \label{fig:f2_visuals}
\end{figure}

\subsubsection{Visual knowledge extraction}

\paragraph{Optical flow} As a component of visual knowledge, optical flow was integrated to provide motion data, which is crucial for detecting moving objects, such as cars. We utilized the Lucas-Kanade~\cite{lucas1981iterative} method for precise optical flow estimation, which enhances feature point tracking across consecutive frames. This method operates under the assumption that flow remains consistent in a pixel's local vicinity, solving optical flow equations for each pixel in this area. Our qualitative analysis revealed that employing the average of optical flows yields more effective visual cues for the VLM. This visually implies a shift from multiple arrows depicting individual flows to a single averaged arrow, supposedly offering a more concise and informative representation of an object's movement as shown in Fig.~\ref{fig:f2_visuals} (d).

\mybox{\textcolor{white}{Instruction Prompt}}{black!40}{black!10}{You are given an image of a crosswalk, sidewalk and road. Image consists of different viewpoints of a walker. The top of the image shows the front angle of the walker,  in the bottom left shows the left angle, the image in the bottom center shows the bottom angle, and the image in the bottom right displays the right angle. Your task is to assess the safeness of crossing street intersections based on image. Please make sure you read and understand these instructions carefully. You should provide a score of safety and reason why.}

\myboxcot{Auto Chain-of-Thought for Risk Assessment}{blue!30}{blue!10}{
1. Check if the cars on the road are approaching the crosswalk or just stopped. 
Check the sign of parallel traffic light of crosswalk

2. Check if the pedestrian signal shows a crossing signal or not.

3. Check if there are any crossing pedestrians around the crosswalk.

4. Repeat this to the whole angles (front, left, bottom, right) of the image.

5. Assign a score for safety on a scale of 1 to 5, where higher score means it is safer to cross the road.
}

\paragraph{Segmentation mask} We employed YOLOv8-Seg, a variant of YOLOv8, to produce segmentation masks for input images. This model incorporates CSPDarknet53~\cite{bochkovskiy2020yolov4} for feature extraction, followed by a C2f module and dual segmentation heads. YOLOv8-Seg was selected for its high-speed performance and accuracy in instance segmentation tasks. A value of 0.7 was used for the \texttt{iou} hyperparameter. For confidence scores, 0.5 was used for the front viewpoints while 0.25 was used for the other viewpoints.

\myboxcriteria{Evaluation Criteria}{green!40}{green!10}{Safety level (-2 to 2) - An assessment of whether the current situation is safe for pedestrians to cross the crosswalk. Each score means the following status:

Score -2 - Totally dangerous: Any car coming toward the crosswalk.

Score -1 - Partially dangerous: There is no car approaching the crosswalk, but the pedestrian light is at a stop sign.

Score 0 - Keep caution: Traffic light is red, yellow or not visible and pedestrian light is a crossing sign.

Score 1 - Partially safe: you can cross the road because there are no cars, traffic light is red, yellow or not visible, the pedestrian light is a crossing sign or not visible, and pedestrians are crossing. Or, there are no cars, the traffic light is green, the pedestrian light is a crossing sign or not visible.

Score 2 - Totally safe: you can cross the road because there are no cars, the traffic light is green, the pedestrian light is crossing sign or not visible and other pedestrian traffic. 
}

\paragraph{Bounding box} We utilized YOLOv8~\cite{Jocher_YOLO_by_Ultralytics_2023} to generate bounding boxes with two pixel coordinates identifying objects within a scene. The class labels and confidence score are depicted in the output to provide additional visual information to the VLM as shown in Fig.~\ref{fig:f2_visuals} (b). However, initial qualitative assessments revealed that YOLOv8, pretrained on the COCO~\cite{lin2014microsoft} dataset encompassing 80 classes, was inadequate for recognizing objects from a robot's egocentric perspective in sidewalk environments. To address this, we finetuned YOLOv8 on the SideGuide~\cite{park2020sideguide} dataset which significantly improved the model's ability to discern features unique to sidewalk environments, leading to enhanced performance of object detection for the VLM. Hyperparameters of 0.7 and 0.25 were used for the values of intersection over union \texttt{(iou)} and minimum confidence threshold \texttt{(conf)}\footnote{https://docs.ultralytics.com/usage/cfg}.

\subsection{Vision-Language Model for Risk Assessment} 
\label{sec:safetyscore}
Motivated by prior successful endeavors in generating automatic metrics using VLMs \cite{Min2023FActScoreFA, Liu2023GEvalNE}, we employ GPT-4V to automatically measure safety scores with inputs of both visual knowledge and text. Note that our approach is based on zero-shot prompting and does not require labeled data to measure scores. Our approach is a prompt-based evaluator with three key components: (1) a prompt including the definition of the task and input details,
(2) a series of intermediate instructions designed to solve problems by outlining detailed evaluation steps,
and (3) a prompt for predicting a safety score based on the given criteria.

\subsubsection{Instruction prompt for safety evaluation}
A prompt is a natural language instruction that outlines a task, providing details on inputs, desired outputs, evaluation criteria, and reasoning verification. In our prompt, we initially present task specifications and instructions to obtain both the safety score and the corresponding reasoning. Moreover, to leverage multiview information, we incorporate explanations of distinct information based on the image patch location as it is important to augment instructions when using additional feature inputs (see Fig.~\ref{fig:prompt_instruction}).

\subsubsection{Auto Chain-of-Thought for safety measurement}
Chain-of-Thought (CoT) is a sequential set of intermediate steps designed to solve a problem systematically~\cite{Wei2022ChainOT}. To enhance reasoning capabilities, we incorporate the Auto CoT~\cite{Liu2023GEvalNE}, which furnishes intermediate instructions to measure scores. In the context of vision-language models and their applications, it demonstrates superior performance, particularly in explaining the evaluation process. Fig.~\ref{prompt:cot} illustrates one of our Auto CoT prompts.

\begin{figure}
    \centering
    \includegraphics[width=.9\linewidth]{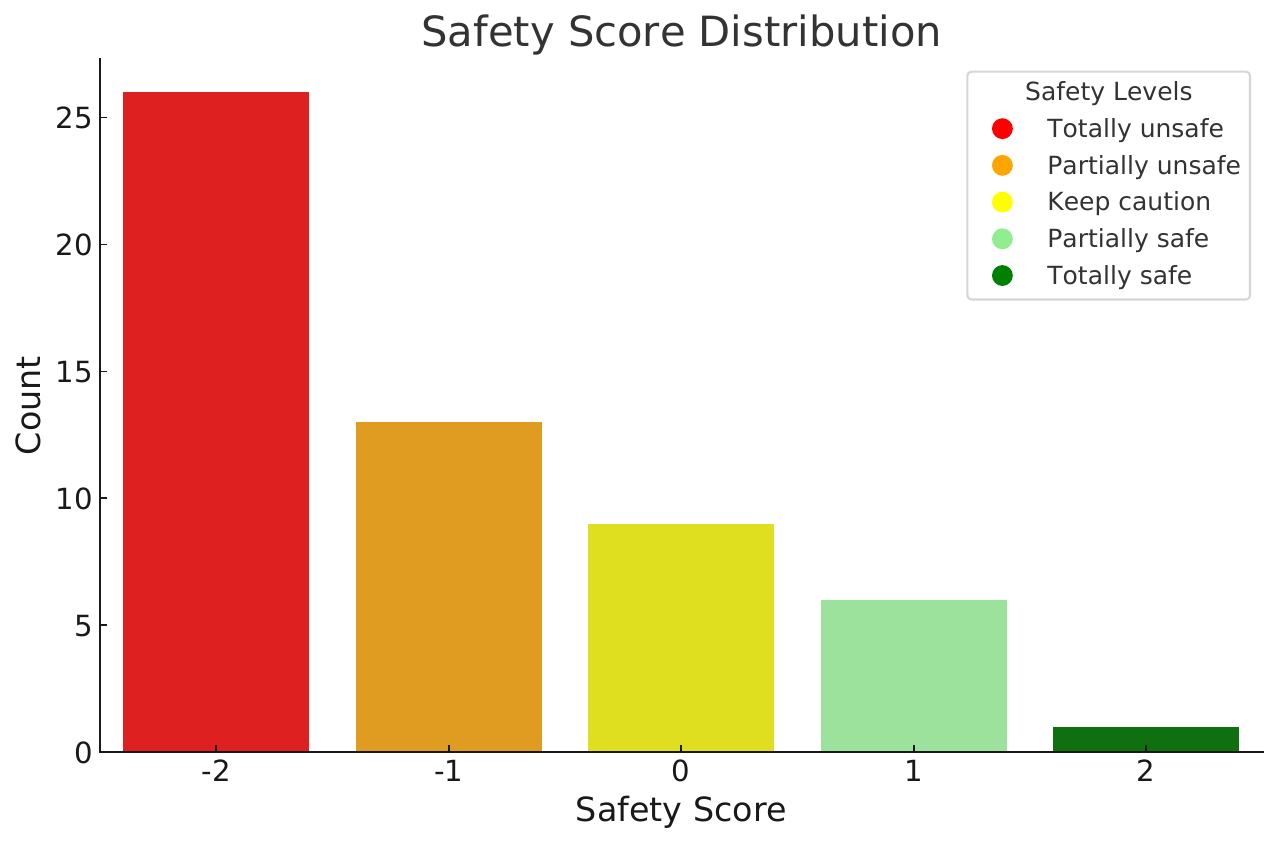}
    \caption{\textbf{Safety score distribution of annotated data.} A large portion of the data contains unsafe scenarios.}
    \label{fig:data_stat}
\end{figure}

\subsubsection{Evaluating safety with criteria and scoring function}
To obtain a desired output, the prompt should contain detailed criteria for measurement. Based on the predefined criteria (Table \ref{tab:safety}), we provide the safety score and description for the safety levels in natural language as shown in Fig.~\ref{prompt:criteria}.

We provide a multiview image $I$ and its extracted visual knowledge of bounding box $I_{bbox}$, optical flow $I_{flow}$, and segmentation mask $I_{mask}$ to the VLM to get a safety score. We develop a reason-verification prompt $E$ that instructs the VLM to measure the safety score $s$ with specified criteria and to verify the reason. Concretely, we provide the image $I$ with its visual knowledge and feed the prompt $E$ to measure the safety score,
$s = VLM(E, I, I_{bbox | flow | mask | none})$.
As a baseline, we use the $I_{none}$ and $E$ prompts, which represent raw images without visual knowledge and text prompts without the Auto CoT prompt, respectively.

\section{EXPERIMENTAL RESULTS}
\label{sec:experiment}
In this section, we evaluate the effectiveness of GPT-4V on real-world street crossing data. Our approach marks a pioneering effort in autonomously calculating safety scores by integrating vision and language in a zero-shot scheme.

\begin{table}[t]
    \centering
    \begin{threeparttable}
    \begin{tabular}{c|cc}
        \toprule 
        \textbf{Method} & \textbf{Accuracy} &  \textbf{Spearman's $\rho$}  \\
        \hline\hline
        Baseline & 0.2545 & \underline{0.3602} \\
        \hline
        + CoT\tnote{1} & 0.1636 & 0.2532 \\
        + bbx & \underline{0.3272}\tnote{2} & 0.1214 \\
        + mask &  \textbf{0.4545} & 0.2757  \\
        + flow & \textbf{0.4545} & \textbf{0.4348} \\
        \bottomrule
    \end{tabular}
    \caption{Model evaluation results.}
    \label{tab:experimental_results}
    \begin{tablenotes}
    \item[1] Chain-of-Thought. 
    \item[2] The \textbf{best} and \underline{second best} performances.
    \end{tablenotes}

    \end{threeparttable}
\end{table}

\subsection{Experiment Setup}
\paragraph{Dataset}
Due to the lack of publicly available datasets on safety metrics, we generated a unique dataset comprising a subset of 55 images collected via a quadruped robot. Although our framework does not require labeled data, we labeled the safety score based on the aforementioned criteria to verify performance. Three researchers independently annotated the data and selected the majority vote for consensus. In cases where no majority was reached, the median annotation was chosen. The Fleiss' $\kappa$ was calculated to be 0.61, indicating substantial agreement among raters. Fig.~\ref{fig:data_stat} illustrates the annotated safety score distribution.

\paragraph{Implementation details} We use GPT-4V to measure the safety score, with the temperature set to $1$ for answer generation. We conducted a series of experiments by varying the composition of prompts and types of visual information to assess their individual effects on safety score prediction.

\paragraph{Evaluation metric} We use accuracy to measure the performance of safety score prediction as mentioned in Sec.~\ref{sec:safetyscore}. However, as the distribution of labels is skewed as shown in Fig.~\ref{fig:data_stat}, we additionally verify with the Spearman's $\rho$ correlation of the output and score labels. 


\begin{table*}[ht]
\centering
\begin{tabular}{m{4.5cm}|m{5.5cm}|m{5.5cm}}
\toprule
\textbf{Images} & \textbf{Baseline} & \textbf{ + Optical Flow} \\ \midrule
\RaiseImage[width=4.5cm, height=4.5cm]{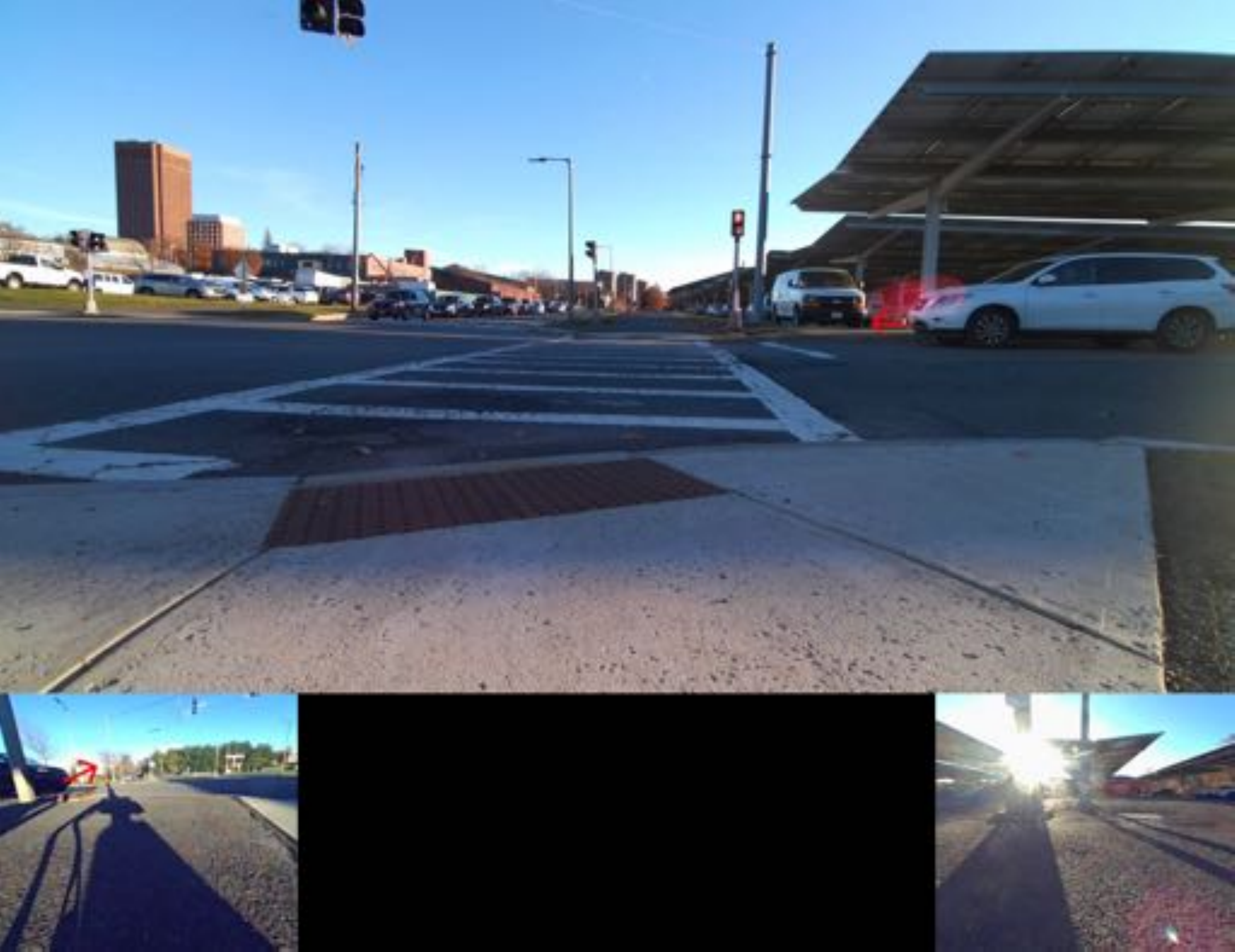} & 
\textbf{There is no car approaching the crosswalk directly}, but the pedestrian light is showing a red hand, indicating a stop signal for pedestrians. &
\textbf{There is a vehicle visibly approaching the crosswalk as indicated by the red arrow showing average optical flow towards the crosswalk area}.
\\ 
\midrule
\RaiseImage[width=4.5cm, height=4.5cm]{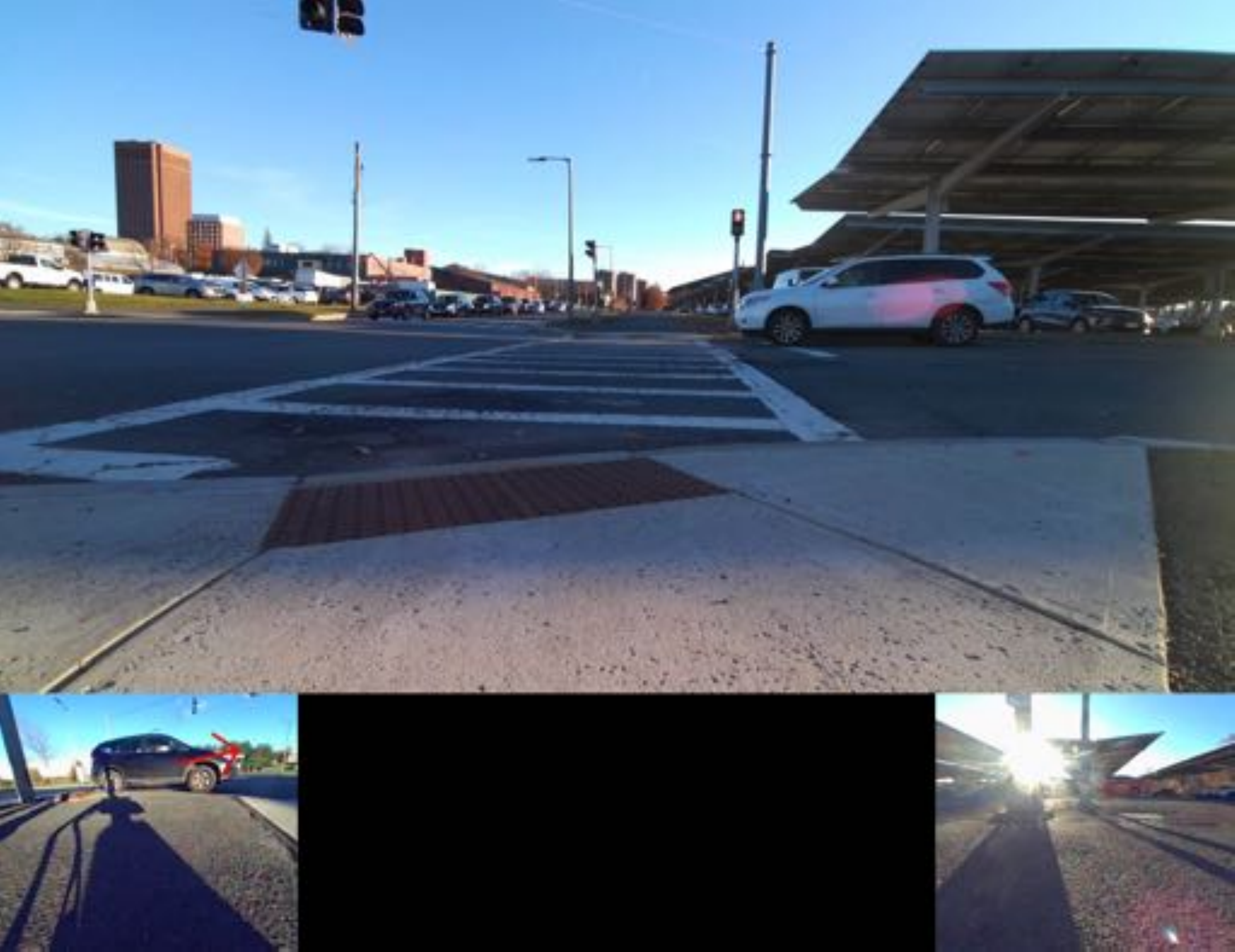} & 
\textbf{There is no car approaching the crosswalk directly}, but the pedestrian signal is red, indicating a stop sign for pedestrians, which makes it partially dangerous to cross.
&
\textbf{There is a vehicle shown in the bottom left image with an arrow indicating its movement toward the crosswalk}, which suggests it is unsafe for pedestrians to cross at this moment
\\
\bottomrule
\end{tabular}
\caption{Qualitative results on scene description} 
\label{tab:optical_flow_qualitative}
\end{table*}

\subsection{Evaluation Results}
Table~\ref{tab:experimental_results} comprehensively outlines the results of various experiments conducted under diverse settings. Initially, the baseline setting is introduced, where the model's performance metrics are quantified in terms of accuracy and correlation degree, with values of 0.2545 and 0.3602 respectively. Upon comparison, the Chain of Thought (CoT) approach demonstrated inferior results, lagging behind the baseline in both accuracy and correlation degree.

Further analysis reveals interesting insights regarding the bounding box information. While it offers a marginal improvement in accuracy, it concurrently leads to a significant reduction in correlation. This discrepancy poses a challenge in drawing definitive conclusions regarding the performance increase.

\begin{figure}
    \centering
    \includegraphics[width=.9\linewidth]{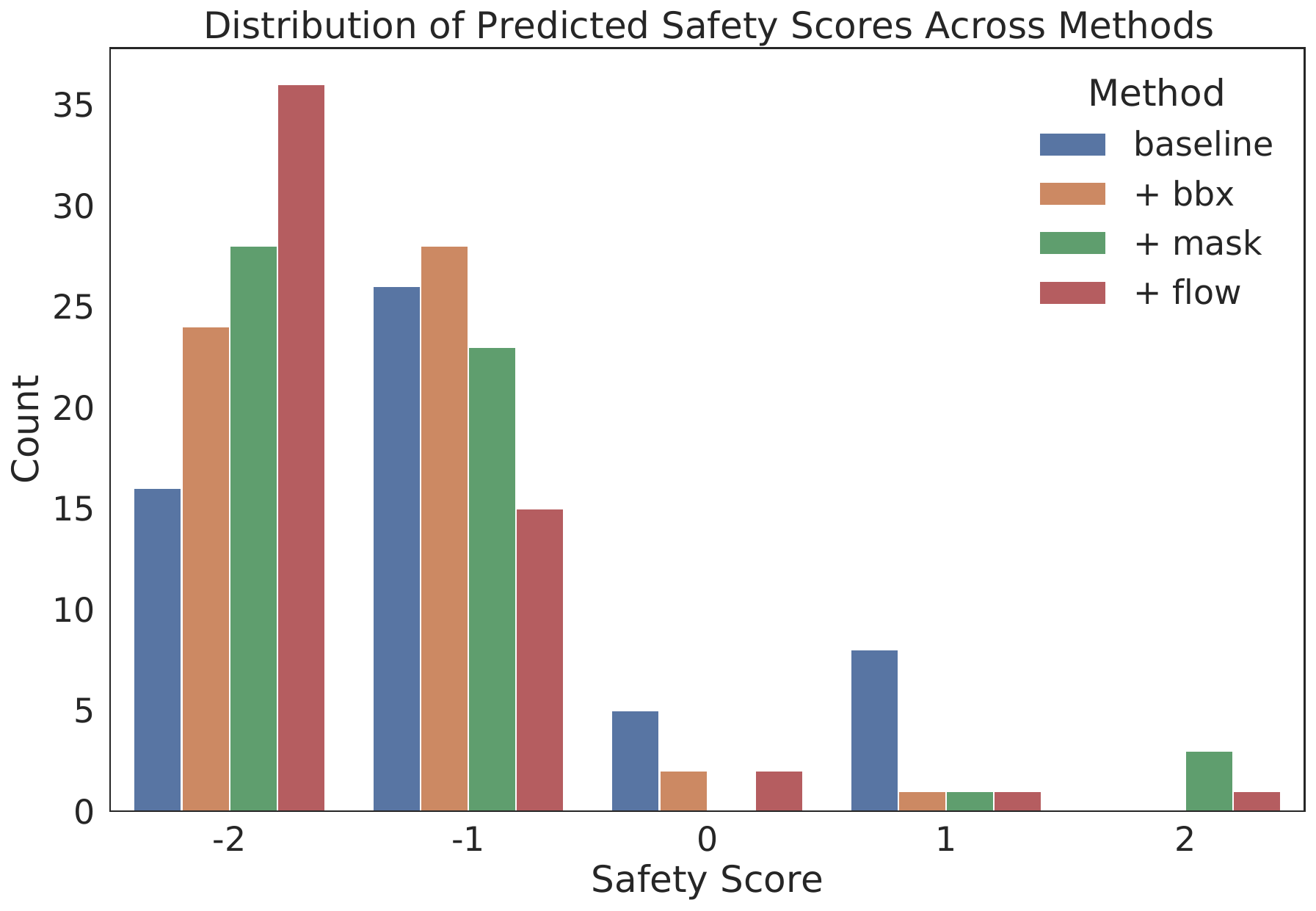}
    \caption{\textbf{Safety score distribution of predictions.} The distribution of safety score predictions from different methods is illustrated with each represented by a unique color.}
    \label{fig:output_label_nums}
\end{figure}

In a contrasting scenario, the segmentation mask feature significantly elevates the accuracy from the baseline's 0.2545 to an impressive 0.4545. However, this improvement comes with a trade-off, as the correlation degree experiences a substantial decrease, dropping from 0.3602 to 0.2757. 

The optical flow information stands out in the experimental results. It not only achieves the highest accuracy among all tested features but also maintains a relatively high correlation degree. This may underscore the significant role of temporal information, as captured by the optical flow feature, in enhancing the suitability and effectiveness for the targeted task. The analysis of these diverse experimental settings provides a nuanced understanding of how different features and methodologies impact the overall performance in task-specific scenarios.




\section{DISCUSSION}
\label{sec:analysis}

\subsection{The Effect of Visual Knowledge}

\paragraph{Bounding Box}
The analysis reveals that the bounding box feature exhibits relatively low correlation degree and acts as noise when compared to the baseline. This is because the information that can be extracted from object detection is already sufficiently captured by the VLM. Moreover, the bounding boxes often obscure important information, such as traffic signals, which complicates the model's ability to make accurate judgments.

\paragraph{Segmentation Mask}


The inclusion of segmentation mask feature has enhanced accuracy while diminishing the correlation degree. This phenomenon correlates with the data distribution observed in Fig.~\ref{fig:data_stat}. Our analysis determined that the model becomes more responsive to risk factors upon receiving a segmentation mask as input, leading to a higher classification of scenes into unsafe categories, as depicted in Fig.~\ref{fig:output_label_nums}. Thus, accuracy improves because the data is skewed to unsafe cases, resulting in a reduced correlation degree.

\paragraph{Optical Flow}
Among visual features, optical flow has shown the highest performance improvement in terms of accuracy. This is a reasonable result considering the impact of the direction and speed of pedestrians and vehicles on safety due to the road environment. In particular, as can be seen in Table~\ref{tab:optical_flow_qualitative}, using the optical flow feature, the risk can be accurately analyzed by recognizing vehicles turning right toward a crosswalk.  This result supports our assumption that temporal information in the scene, especially vehicle direction information, is an important factor in assessing risk. Furthermore, the correlation degree is observed to be higher than the baseline unlike other visual features.

\subsection{Does Chain-of-Thought Really Help?}
In contrast to the findings of Liu et al.~\cite{Liu2023GEvalNE}, the incorporation of CoT into the prompt did not enhance performance in our task; rather, it resulted in degradation. We hypothesize two potential reasons for this outcome after analyzing the reasoning outputs.

First, the reasoning produced through our method is notably more verbose than that of the baseline, displaying excessive sensitivity to keywords embedded within the CoT instructions. For example, while the baseline typically generates concise explanations comprising two to three sentences, the responses from CoT-informed prompts extend beyond three sentences. Moreover, in scenarios involving a car positioned a considerable distance from the crosswalk, the focus tends to be on the mere existence of the car, leading to an interpretation of an unsafe situation.

Second, our CoT framework might not be optimally configured for the task at hand. The effectiveness of the generated content relies on the specifics of the prompt, suggesting that our current CoT approach may not be ideally tailored for assessing safety. Furthermore, despite the proven success of CoT methodologies in NLP tasks, their applicability and performance in interpreting multimodal data have yet to be conclusively established.
\section{CONCLUSIONS}
\label{sec:conclusion}

We evaluate a vision-language model for safety-aware scene understanding in robot navigation, particularly to support BLV individuals to safely cross streets. By leveraging GPT-4V, we evaluate the capabilities of a comprehensive and nuanced understanding of street crossing environments, aiming for the development of a reliable mobility aid. We share experimental results that provide valuable insights into the effects of various visual features. Optical flow emerged as a significant feature, underscoring the importance of temporal information in enhancing performance. 

While our findings underscore the effectiveness of the VLM in facilitating safety-aware decisions for street crossing, this is not without its limitations. Notably, the application of the Chain-of-Thought methodology did not yield the anticipated performance improvements. This discrepancy highlights the need for further refinement in prompt engineering and adaptation strategies for visual data. Furthermore, our dataset's sensitivity to specific risk factors and its tendency to overestimate risk in certain scenarios highlight the need for greater diversity and quantity in data collection.

Future research will focus on several key areas: investigating the impact of different viewpoints on model performance, exploring the importance of temporal understanding in navigation decisions, comparing the performance of our model with vision-only models, increasing inference speed, and fine-tuning the model based on user-specific preferences. Addressing these areas is crucial for advancing our goal of developing a reliable mobility aid that ultimately supports the mobility of blind and low-vision (BLV) individuals.

\addtolength{\textheight}{-12cm}   





\bibliographystyle{include/IEEEtran} 
\bibliography{include/IEEEabrv,include/root}

\end{document}